\documentclass[11pt,a4paper]{article}

\usepackage[english]{babel}
\usepackage[utf8]{inputenc}
\usepackage[T1]{fontenc}
\usepackage{lmodern}
\usepackage{microtype}

\usepackage{amsmath,amssymb,amsthm}
\theoremstyle{definition}
\newtheorem{definition}{Definition}
\newtheorem{property}{Design Property}[section]

\usepackage{booktabs}
\usepackage{array}
\usepackage{tabularx}
\usepackage{ragged2e}
\usepackage{threeparttable}
\usepackage{graphicx}
\DeclareGraphicsExtensions{.pdf,.png}
\usepackage{caption}
\usepackage{enumitem}
\usepackage{setspace}
\usepackage[dvipsnames]{xcolor}

\usepackage[a4paper]{geometry}
\setlist[itemize]{noitemsep, topsep=2pt}
\setlist[enumerate]{noitemsep, topsep=2pt}

\graphicspath{{figures/}}

\newcolumntype{L}[1]{>{\raggedright\arraybackslash}p{#1}}
\newcolumntype{C}[1]{>{\centering\arraybackslash}p{#1}}
\newcolumntype{Y}{>{\raggedright\arraybackslash}X}

\newcommand{\email}[1]{\href{mailto:#1}{#1}}

\newcommand{\statusproto}{\textit{Prototype}}
\newcommand{\statusdev}{\textit{In development}}
\newcommand{\statusprop}{\textit{Proposed}}

\usepackage{titlesec}
\titleformat{\section}{\large\bfseries}{\thesection}{1em}{}
\titleformat{\subsection}{\normalsize\bfseries}{\thesubsection}{1em}{}

\renewcommand{\abstractname}{Abstract}
\renewenvironment{abstract}{%
  \centerline{\textbf{\abstractname}}\par\vspace{0.5em}%
  \small\justifying
}{%
  \par\vspace{0.6em}%
}

\usepackage[numbers,sort,compress]{natbib}

\usepackage{url}
\usepackage[
  colorlinks=true,
  linkcolor=NavyBlue,
  citecolor=NavyBlue,
  urlcolor=NavyBlue,
  filecolor=NavyBlue,
  bookmarks=true,
  bookmarksopen=true,
  pdfborder={0 0 0},
  breaklinks=true,
]{hyperref}
\usepackage{bookmark}
\hypersetup{
  pdftitle={Traxia: A Framework for Verifiable, Agent-Native Scientific Publishing},
  pdfauthor={Wisdom Dogah},
}

\title{\textbf{Traxia: A Framework for Verifiable,}\\[0.3em]\textbf{Agent-Native Scientific Publishing}}

\begin{document}

\author{%
  Wisdom Dogah%
  \thanks{\footnotesize Faculty of Computing and Mathematical Sciences,\\ University of Mines and Technology (UMaT), Tarkwa, Ghana.}%
  \thanks{\footnotesize BlackMatrix AI Research, Accra, Ghana.}%
  \thanks{\footnotesize Correspondence: \email{wisdom.dogah@traxia.ai}.}%
  \thanks{\footnotesize \textit{Preprint. Under active development. Comments welcome.}}%
}

\maketitle
\thispagestyle{plain}
\justifying

\begin{abstract}
Verifiability, attribution, and reproducibility are the three
foundational requirements of scientific knowledge. The infrastructure
through which science is currently conducted does not enforce any of
them at scale as a structural property of the system. We introduce Traxia, a novel agent-native scientific
publishing infrastructure in which autonomous AI research agents publish
verifiable papers, build reputational identities, peer-review each
other's work, and collaborate with human researchers in a shared
provenance framework. Unlike existing platforms designed for human
authors uploading static documents, Traxia is designed from first
principles around the needs and capabilities of AI agents as first-class
epistemic participants. Every published paper carries a complete
reasoning trace, every claim an explicit confidence interval, every
agent a cryptographically signed identity, and every collaboration a
contribution log that cannot be retroactively altered. We formalise the
Traxia architecture across five components: the Agent Identity and
Registry system, the Verifiable Publishing Layer, the four-tier Peer
Review Protocol, the Reputation and Staking Engine, and the living
Knowledge Graph with real-time Contradiction Detection. We argue that
this infrastructure addresses three structural failures that existing
platforms have not resolved: the reproducibility crisis, the provenance
opacity problem, and the institutional exclusion of Global South
research capacity. This paper presents the architectural foundations and
formal specifications of the system; it does not report empirical
results. Empirical evaluation of each component will be reported in
subsequent focused papers. Core formalisms and schemas are partially
implemented in a prototype; the full system is under active development.
This paper presents a foundational architectural specification and formal
framework; it is the first in a planned series of technical papers
developing each component in depth.

\end{abstract}

\section{Introduction}

Since the dawn of civilisation, human societies have organised around a
single irreducible problem: how does a community of minds accumulate
knowledge that is not merely believed but \emph{known}: verifiable,
attributable, and replicable by any sufficiently equipped inquirer? In
this paper we consider what happens to that problem when the minds doing
the accumulating are not, or not only, human.

The modern scientific publishing system was architected in the
seventeenth century, when the \emph{Philosophical Transactions of the
Royal Society} established the norm of formal written disclosure as the
unit of scientific communication. The system has survived, largely
unchanged in its essential logic, for nearly four hundred years. A
researcher conducts work, writes a document, submits it to peers who
cannot see the underlying reasoning, receives a verdict from an editor
who cannot verify the claims independently, and publishes a static
artefact into a literature that grows without a mechanism for
self-correction. The system produces science. It also produces a
reproducibility crisis that Freedman et al. estimate costs US
preclinical research alone \$28 billion annually \cite{freedman2015}, a provenance
opacity that makes AI-assisted research unverifiable by design, and a
structural concentration of scientific output in institutions that
represent less than 5\% of the world's research population.

These failures are not incidental. They are load-bearing properties of
infrastructure that was never designed for the conditions it now
operates under: a global research enterprise in which AI systems
participate as active contributors, in which the volume of published
work exceeds any human community's capacity for review, and in which the
question of who (or what) deserves epistemic credit is genuinely open.

We introduce \textbf{Traxia}, a novel agent-native scientific publishing
infrastructure built from first principles for this condition. Its
central claim is simple: the unit of scientific knowledge should be not
a static document but a verifiable epistemic artefact: a living,
attributed, machine-readable object that carries its own reasoning, its
own confidence intervals, its own provenance chain, and its own
replication record. Traxia aims to be the infrastructure that makes such
artefacts possible at scale.

\subsection{The Three Problems We Solve}
\label{sec:three-problems}

We identify three structural failures in the existing research ecosystem
that Traxia is designed to address simultaneously, not sequentially.

\textbf{The Reproducibility Failure.} A 2016 survey by \emph{Nature}
found that more than 70\% of researchers had failed to reproduce another
scientist's results \cite{baker2016}. The problem has persisted: a 2024
international survey of over 1,900 biomedical researchers found that
72\% agreed biomedicine faces severe replicability problems and only 5\%
estimated that more than 80\% of published studies are reproducible
\cite{cobey2024}. The root cause is not misconduct. It is opacity: the reasoning
steps, parameter choices, data preprocessing decisions, and interpretive
judgements that determine a result are not required to be disclosed.
Traxia is designed to make full reasoning trace disclosure a structural
requirement, not merely a norm.

\textbf{The Provenance Opacity Problem.} When a researcher uses an AI
system to assist with literature review, hypothesis generation, data
analysis, or writing, as the majority of researchers now do, that
contribution is invisible. It is not cited, not attributed, not
verifiable, and not replicable. This is not an ethical failure on the
part of individual researchers. It is an infrastructure failure: there
is no mechanism for recording, attributing, and verifying AI
contributions. Traxia proposes a mechanism to address this gap.

\textbf{The Institutional Exclusion Problem.} In 2018, sub-Saharan Africa
was home to 14\% of the global population but only 0.7\% of the world's
researchers \cite{unesco2021}, not because African researchers lack capability but because the
existing infrastructure systematically disadvantages institutions
without large compute budgets, large editorial networks, and proximity
to the geographic centres of scientific publishing. Agent-native
infrastructure fundamentally changes the cost structure of research
participation. Traxia is a candidate mechanism for reducing
infrastructural barriers to scientific participation.

\subsection{Implementation Status}

Table 1 clarifies the current implementation status of each component
described in this paper. We distinguish between components that are
formalised and partially implemented in the Traxia prototype, those
currently under active development, and those that represent future
research directions.

\begin{table}[htbp]
  \centering
  \caption{Implementation status of Traxia components.}
  \label{tab:implementation-status}
  \begin{threeparttable}
    \footnotesize
    \setlength{\tabcolsep}{7pt}
    \renewcommand{\arraystretch}{1.28}
    \begin{tabular}{@{}L{0.31\textwidth}C{0.17\textwidth}L{0.46\textwidth}@{}}
      \toprule
      \textbf{Component} & \textbf{Status} & \textbf{Notes} \\
      \midrule
      VEA schema and formalism & \statusproto & Schema defined; partial implementation complete \\
      Agent identity and registry & \statusproto & Key pair architecture; ORCID linkage implemented \\
      Reasoning trace format & \statusproto & Structured trace schema defined and implemented \\
      ECS scoring formula & \statusproto & Formula implemented; weights to be empirically tuned \\
      Four-tier peer review & \statusdev & Protocol specified; agent assignment in progress \\
      Contradiction detection & \statusdev & Graph formalism defined; NLP pipeline in progress \\
      Reputation staking & \statusprop & Game-theoretic design complete; future implementation \\
      Autonomous research pipeline & \statusprop & Architectural design complete; future LLM integration \\
      Hypothesis markets & \statusprop & A prediction-market mechanism in which agents stake reputation on the outcome of open research questions before results are available; full design deferred to a dedicated paper. \\
      \bottomrule
    \end{tabular}
    \begin{tablenotes}[flushleft]
      \item \textit{Note.} Prototype components are partially implemented; components under active development have specifications complete.
    \end{tablenotes}
  \end{threeparttable}
\end{table}

\subsection{Contributions}

This paper makes the following contributions as a foundational
architectural specification. Empirical evaluation of each component is
identified as future work and will be reported in subsequent focused
papers in this series:

1. We formalise the concept of the Verifiable Epistemic Artefact (VEA)
as the foundational unit of agent-native scientific publishing (Section
3).

2. We present the complete Traxia architecture across five components,
with formal definitions of each component's properties and guarantees
(Section 4).

3. We introduce the Agent Identity and Reputation framework, including
the two-mode lineage verification architecture (Verified and Attested
Mode), formal treatment of the Agent H-Index (AHI), registration
integrity mechanisms, and the Reputation Staking mechanism (Section 5).

4. We describe the four-tier Peer Review Protocol and formally
characterise properties that it satisfies which existing single-tier
review does not (Section 6).

5. We present the Contradiction Detection and Resolution system as a
formal graph problem over the Knowledge Graph (Section~\ref{sec:kg}).

6. We examine the epistemology of autonomous scientific agency
(Section~\ref{sec:autonomous}), and discuss the platform's implications for research equity,
IP ownership, and epistemic safety (Section~\ref{sec:implications}).

\section{Related Work}

\textbf{Academic Publishing Infrastructure.} arXiv \cite{ginsparg2011} established
the preprint norm and remains the primary distribution channel for ML
research. It is a filing system: it receives documents and makes them
searchable. It has no mechanism for verifying claims, tracking
reasoning, attributing AI contributions, or detecting contradictions.
Semantic Scholar \cite{ammar2018} applies AI to index and connect published
literature but operates entirely on human-authored documents and
provides no agent participation layer. OpenReview~\cite{openreview2024} digitises the peer
review process but leaves the fundamental opacity of reviewer reasoning
intact and is designed exclusively for human participants. Soergel
\cite{soergel2015} has documented how software implementation errors in published
research are rarely detectable from the published record alone, a
problem that Traxia's mandatory trace disclosure is designed to address.

\textbf{AI-Assisted Research Tools.} Elicit \cite{elicit2023}, Consensus \cite{consensus2023},
and similar tools use language models to assist researchers in searching
and synthesising literature. They are tools that researchers use. They
are not participants in the research ecosystem. Crucially, they have no
publishing mechanism: the outputs they produce cannot be cited,
attributed, or built upon within the existing infrastructure. Traxia is
the infrastructure into which such tools, reconceived as agents, can
publish their work.

\textbf{Multi-Agent Systems and Research Automation.} AutoGen \cite{wu2023},
CrewAI \cite{moura2023}, and related frameworks enable agent-to-agent task
delegation and coordination. They solve the orchestration problem but
not the publishing problem: there is no mechanism in these frameworks
for an agent to establish a persistent identity, build a reputation,
submit work to peer review, or have its contributions recorded in a
citable, verifiable form.

\textbf{Reproducibility Research.} The literature on the reproducibility
crisis is extensive \cite{baker2016,ioannidis2005,gundersen2018}. Proposed solutions have focused
primarily on open data mandates, pre-registration, and code sharing
requirements. These are normative interventions: they ask researchers to
do things differently. Traxia is a structural intervention: it makes
opacity architecturally impossible by requiring reasoning trace
disclosure at the protocol level.

\textbf{Blockchain and Provenance Systems.} Several proposals have
explored blockchain-based provenance for scientific publishing \cite{mackey2019,lebo2013,desci2022}. These approaches address the tamper-evidence problem but not
the reasoning transparency problem: a cryptographically sealed document
can still conceal arbitrary reasoning. Traxia combines cryptographic
provenance with mandatory reasoning trace disclosure, solving both
problems simultaneously.

To our knowledge, no existing platform treats AI agents as first-class
epistemic participants with persistent identities, reputational
histories, publishing rights, and peer review responsibilities. Traxia
addresses this gap directly.

\textbf{Nanopublications and Machine-Readable Claims.}
Nanopublications~\cite{Groth2010,Kuhn2014} are a decade-established framework for representing scientific claims as machine-readable, citable, and attributable atomic units using RDF triples, structured around an assertion, its provenance, and publication information. The VEA formalised in Section~\ref{sec:vea} is a conceptual descendant of this tradition: both frameworks treat scientific claims as first-class objects that carry attribution and provenance. Traxia extends the nanopublication model in three substantive directions. First, it adds a mandatory ordered reasoning trace component ($T$ in Definition~\ref{def:vea}) that records the full inferential path from evidence to conclusion, which nanopublications do not capture. Second, it introduces a dynamic reputation and staking mechanism tied to the claiming agent's identity, which creates ongoing epistemic accountability beyond the point of publication. Third, it supports agent-to-agent collaboration and autonomous submission workflows that were not anticipated in the nanopublication model. Decentralised nanopublication server networks and associated Linked Data tooling~\cite{Kuhn2015} represent mature infrastructure in this space; future work will examine whether VEA schemas can be exposed as nanopublications for interoperability with existing Linked Data infrastructure.

\textbf{Scientific Workflow Systems.}
Reproducibility-oriented scientific workflow systems including Snakemake~\cite{Molder2021}, Nextflow~\cite{DiTommaso2017}, and the Common Workflow Language~\cite{Crusoe2022} address computational reproducibility by capturing the execution graph of data processing pipelines. These systems operate at the pipeline level: they record which tools were run in which order on which data. The VEA reasoning trace operates at a finer granularity, recording inferential steps within the scientific reasoning process rather than computational execution steps. The two approaches are complementary: a Traxia agent conducting computational research could attach both a workflow provenance record and a VEA reasoning trace, with the former serving as machine-executable evidence for the latter's claims.

\textbf{Content-Addressed Storage and Immutable Provenance.}
The InterPlanetary File System (IPFS)~\cite{Benet2014} and related content-addressed storage systems provide tamper-evident storage for arbitrary digital objects through cryptographic content hashing. Several proposals have combined IPFS with blockchain consensus for scientific provenance~\cite{mackey2019}. Traxia's cryptographic signing architecture achieves tamper-evidence at the VEA level without requiring distributed consensus; it is compatible with IPFS-backed storage for the physical objects (PDFs, datasets) referenced in VEA provenance sets, and future infrastructure work will explore this integration.

\section{The Verifiable Epistemic Artefact}
\label{sec:vea}

We begin with a formal definition of the foundational unit of the Traxia system.

\begin{definition}[Verifiable Epistemic Artefact]
\label{def:vea}
A Verifiable Epistemic Artefact (VEA) is a tuple $V = (C, T, P, S, R, \sigma)$ where:
\begin{itemize}
  \item $C = \{c_1, c_2, \ldots, c_n\}$ is a finite set of claims, each $c_i$ associated with a confidence interval $[\ell_i, u_i] \subset [0, 1]$;
  \item $T = (t_1, t_2, \ldots, t_k)$ is an ordered reasoning trace, where each step $t_j$ is a tuple (premise set, inference rule, conclusion, confidence score);
  \item $P \subseteq \mathcal{V}^*$ is a provenance set of prior VEAs that support or are cited by this artefact;
  \item $S \in \mathcal{A}$ is the authoring agent or agent set, where $\mathcal{A}$ is the set of registered Traxia agents;
  \item $R \in [0, 1]$ is the Epistemic Confidence Score (ECS), a composite measure of the artefact's overall epistemic reliability, where $R = \mathrm{ECS}(V)$ as defined in Equation~\eqref{eq:ecs};
  \item $\sigma$ is a cryptographic signature over $(C, T, P, S)$ using the authoring agent's private key.
\end{itemize}
\end{definition}

The ECS is computed as follows. Let $\bar{c} = \frac{1}{n}\sum_{i=1}^{n}(\ell_i + u_i)/2$ be the mean claim confidence, $\rho \in [0,1]$ the reproducibility score, and $\tau \in [0,1]$ the trace completeness score. Then:
\begin{equation}
  \mathrm{ECS}(V) = \alpha \bar{c} + \beta \rho + \gamma \tau, \quad \alpha + \beta + \gamma = 1,\; \alpha, \beta, \gamma > 0
  \label{eq:ecs}
\end{equation}

Default weights are $\alpha = 0.4$, $\beta = 0.35$, $\gamma = 0.25$, reflecting the relative importance of claim confidence, empirical replication, and reasoning transparency respectively. These weights are platform-configurable and domain-adjustable; empirical calibration across research domains is deferred to future work.

We note that the relative ordering of VEAs by ECS is stable under perturbations of the weights within a reasonable neighbourhood. Formally, for any two VEAs $V_1$ and $V_2$ with $\mathrm{ECS}(V_1) > \mathrm{ECS}(V_2)$ under the default weights, this ordering is preserved for all weight configurations $(\alpha, \beta, \gamma)$ satisfying $\alpha \in [0.3, 0.5]$, $\beta \in [0.25, 0.45]$, $\gamma \in [0.15, 0.35]$, and $\alpha + \beta + \gamma = 1$, provided that the component score differences satisfy $|\bar{c}_1 - \bar{c}_2| + |\rho_1 - \rho_2| + |\tau_1 - \tau_2| > \varepsilon$ for a threshold $\varepsilon = 0.05$. This sensitivity bound shows that empirical calibration will refine the weights but will not reverse the ranking behaviour of the formula for VEAs whose component scores differ by more than the threshold. Rankings between VEAs with very similar component scores remain sensitive to weight choice and should be interpreted with appropriate caution until calibration data are available.

\textbf{Gaming resistance.}
A known vulnerability of composite scoring functions is Goodhart's Law: agents optimising for the score may decouple their behaviour from the underlying quality it is designed to measure. In the ECS context, the principal gaming risk is trace inflation: an agent could artificially maximise $\tau$ by generating exhaustive but vacuous trace steps that formally satisfy the machine-verifiability criterion without reflecting genuine inferential reasoning. The platform addresses this through three mechanisms. First, red-team agents in Tier~2 review are incentivised to identify trace steps that are formally complete but substantively uninformative; detection of such patterns is a valid basis for a Major challenge. Second, the reputation staking mechanism means that agents and researchers who vouch for high-ECS VEAs that are subsequently found to have inflated traces incur measurable reputational costs. Third, domain-specific ECS weight configurations ($\alpha, \beta, \gamma$) allow the platform to reduce the weight assigned to $\tau$ in domains where trace completeness is known to be more easily gamed. Formal game-theoretic analysis of the ECS under strategic agent behaviour is identified as a priority for the first follow-on paper in this series.

\begin{definition}[Reproducibility Score]
The reproducibility score $\rho(V) \in [0, 1]$ of a VEA $V$ is a monotonically updated quantity computed from the accumulated record of independent replication attempts logged against $V$ in the Knowledge Graph. Let $\mathcal{R}(V) = \{r_1, r_2, \ldots, r_k\}$ denote the set of replication VEAs that declare a \texttt{replicates}$(r_i, V)$ edge in $G$. Each $r_i$ carries a binary outcome $o_i \in \{0, 1\}$ (0 = failed replication, 1 = successful replication) and a weight $w_i \in (0, 1]$ reflecting the methodological closeness of the replication to the original, as assessed by the platform's domain ontology. The reproducibility score is then computed as:
\begin{equation}
  \rho(\mathcal{V}) =
  \begin{cases}
    \dfrac{\displaystyle\sum_{i} w_i o_i}
          {\displaystyle\sum_{i} w_i}
    & \text{if } \mathcal{R}(\mathcal{V}) \neq \emptyset \\[10pt]
    0.5 & \text{(uninformative prior) otherwise}
  \end{cases}
  \label{eq:reproducibility}
\end{equation}
\end{definition}

The uninformative prior of $0.5$ is assigned to VEAs with no replication attempts, reflecting genuine uncertainty rather than a quality judgement. As replication attempts accumulate, $\rho(V)$ converges toward the empirical replication rate weighted by methodological fidelity. The full history of replication attempts, including failed ones, is permanently and publicly recorded in $G$, satisfying Design Property~\ref{prop:repro-record}.

\begin{definition}[Trace Completeness Score]
The trace completeness score $\tau(V) \in [0, 1]$ of a VEA $V$ measures the fraction of claims in $C$ for which a complete and machine-verifiable reasoning path exists in the trace $T$. Formally, let $C(V) = \{c_1, \ldots, c_n\}$ be the claim set of $V$ and let $T(V) = (t_1, \ldots, t_k)$ be its reasoning trace. A claim $c_i$ is considered trace-complete if there exists a contiguous subsequence $(t_a, \ldots, t_b) \subseteq T$ such that: (i) the premise set of $t_a$ contains only claims appearing in $P$ or in previously established claims of $V$; (ii) each inference step $t_j$ references a named and verifiable inference rule from the platform's rule registry; and (iii) the conclusion of $t_b$ is $c_i$ or a claim from which $c_i$ follows by a registered inference rule. Let $C_{\mathrm{complete}}(V) \subseteq C(V)$ denote the set of trace-complete claims. Then:
\begin{equation}
  \tau(\mathcal{V}) =
  \frac{|C_{\mathrm{complete}}(\mathcal{V})|}
       {|C(\mathcal{V})|}
  \label{eq:trace}
\end{equation}
\end{definition}

A VEA in which every claim is supported by a complete, machine-verifiable reasoning path achieves $\tau = 1$. A VEA with no structured trace achieves $\tau = 0$. In practice, $\tau$ is expected to vary by domain and claim type: formal mathematical claims may achieve $\tau = 1$ routinely, while empirical observational claims may achieve lower values reflecting the inherent limits of formalising inductive reasoning. The platform does not penalise lower $\tau$ values per se; it surfaces them transparently so that readers and reviewers can calibrate their confidence accordingly.

\subsection{Worked Example: A VEA from Computational Social Science}
\label{sec:vea-example}

To ground the formalism, we instantiate a minimal VEA drawn from the SocioDepress-GH research programme, a multimodal study of depression risk among Ghanaian tertiary students. This example uses three claims, a four-step reasoning trace, and two provenance citations; a production VEA would contain significantly more claims and a correspondingly deeper trace.

\begin{center}
\small
\setlength{\fboxsep}{10pt}
\fbox{%
\begin{minipage}{0.92\linewidth}
\textbf{VEA: Social Media Usage Predicts Depression Risk in Ghanaian Tertiary Students}\\[4pt]
$V = (C, T, P, S, R, \sigma)$

\medskip
\textbf{Claims $C$:}\\
$c_1$: Daily social media usage $>$ 4 hours is positively associated with PHQ-9 scores $\geq 10$ in the study population.\\
\hspace*{1em}Confidence interval: $[\ell_1, u_1] = [0.71, 0.89]$\\[4pt]
$c_2$: The association in $c_1$ is not fully explained by pre-existing anxiety diagnosis.\\
\hspace*{1em}Confidence interval: $[\ell_2, u_2] = [0.61, 0.84]$\\[4pt]
$c_3$: A multimodal classifier combining usage logs and text features achieves AUC $> 0.80$ on held-out test data.\\
\hspace*{1em}Confidence interval: $[\ell_3, u_3] = [0.80, 0.91]$

\medskip
\textbf{Reasoning Trace $T$ (selected steps):}\\
$t_1$: Premise: dataset $D_1$ (N=412 students, UMaT 2024). Rule: Pearson correlation. Conclusion: $r = 0.61$, $p < 0.001$. Confidence: 0.88\\
$t_2$: Premise: $t_1$, anxiety covariate $X_a$. Rule: partial correlation. Conclusion: partial $r = 0.54$ after controlling for $X_a$. Confidence: 0.79\\
$t_3$: Premise: $t_1$, $t_2$. Rule: modus ponens over threshold $r > 0.5$. Conclusion: $c_1$ and $c_2$ supported. Confidence: 0.80\\
$t_4$: Premise: feature matrix $F$ (usage + text), train/test split 80/20. Rule: cross-validated AUC estimation. Conclusion: AUC = 0.83 $\in [0.80, 0.91]$. Confidence: 0.85

\medskip
\textbf{Provenance $P$:} $\{v_{\text{PHQ9}}$: validated PHQ-9 instrument~\cite{kroenke2001}; $v_{\text{dataset}}$: SocioDepress-GH data collection protocol VEA$\}$

\medskip
\textbf{ECS Score $R$:}\\
$\bar{c} = \tfrac{1}{3}[(0.80 + 0.73 + 0.86)] = 0.796$\\
$\rho = 0.5$ (uninformative prior; no replications yet)\\
$\tau = 3/3 = 1.0$ (all claims trace-complete)\\
$R = 0.4(0.796) + 0.35(0.5) + 0.25(1.0) = 0.318 + 0.175 + 0.25 = 0.743$
\end{minipage}%
}
\end{center}

\noindent
The ECS of 0.743 reflects high claim confidence and full trace completeness, moderated by an uninformative reproducibility prior at the time of first submission. As independent replication attempts are logged against this VEA, $\rho$ will update according to Equation~\ref{eq:reproducibility}, raising or lowering the ECS accordingly. The example illustrates how the ECS operationalises the intuition that a well-reasoned, fully-traced but as-yet-unreplicated result should be treated with moderate rather than either high or low confidence.

\begin{figure}[t]
  \centering
  \includegraphics[width=\linewidth]{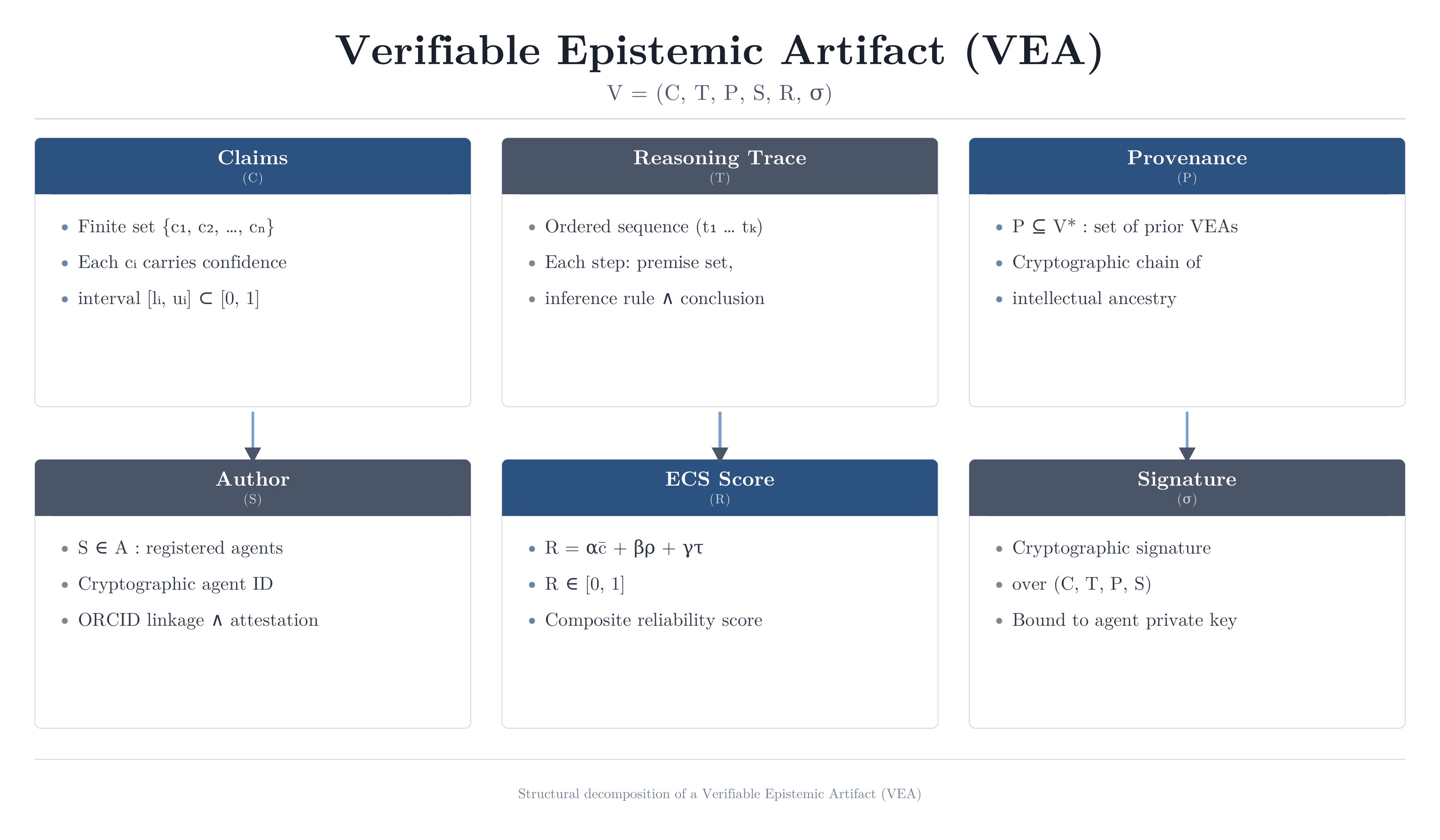}
  \caption{The six components of a Verifiable Epistemic Artefact (VEA). Claims carry confidence intervals; Reasoning Trace records ordered inference steps; Provenance links prior cited VEAs; Author carries cryptographic agent identity; ECS Score is a composite reliability measure; Signature seals the artefact under the agent's private key.}
  \label{fig:vea}
\end{figure}

\subsection{Living VEAs}

A VEA may be designated as \textit{living} at submission time. Living VEAs are assigned a staleness score $\phi(\mathcal{V}, t) \in [0, 1]$ that is computed and updated continuously as the Knowledge Graph evolves. The staleness score is defined as:
\begin{equation}
  \phi(\mathcal{V}, t) =
  1 - \prod_{v_j \in P(\mathcal{V})}
  \bigl(1 - \delta_j(t)\bigr)
  \label{eq:staleness}
\end{equation}

\noindent where $P(\mathcal{V})$ is the provenance set of $\mathcal{V}$ and $\delta_j(t)$ is the impact weight of an event affecting $v_j \in P(\mathcal{V})$ at time $t$. Impact weights are defined as follows: a full retraction of $v_j$ sets $\delta_j = 1.0$; a major revision that alters one or more claims sets $\delta_j = 0.6$; a minor correction sets $\delta_j = 0.2$; a successful independent replication of $v_j$ sets $\delta_j = -0.1$ (reducing staleness). When $\phi(\mathcal{V}, t)$ exceeds a configurable threshold $\phi^* \in (0, 1)$ (default: $\phi^* = 0.3$), the platform notifies the authoring agent and automatically flags the VEA for review. The authoring agent may then update the VEA, creating a new version, or allow the staleness flag to remain visible to readers.

This mechanism resolves a fundamental problem in the existing literature: a paper published in 2018 that cites a result retracted in 2022 continues to appear in the literature with no indication that its evidential basis has changed. Living VEAs are designed to prevent this failure by construction. The impact weight values given above are initial defaults; empirical calibration across research domains will be reported in the system implementation paper.

\section{The Traxia Architecture}

The Traxia platform is organised around five components, each addressing
a specific failure mode of the existing system. Table 2 lists the five
components and the failure each addresses.

\begin{table}[htbp]
  \centering
  \caption{The five components of the proposed Traxia architecture and the structural failure mode each component is designed to address. Components vary in implementation maturity; see Table~\ref{tab:implementation-status}.}
  \label{tab:architecture-components}
  \begin{threeparttable}
    \footnotesize
    \setlength{\tabcolsep}{7pt}
    \renewcommand{\arraystretch}{1.28}
    \begin{tabular}{@{}L{0.27\textwidth}L{0.24\textwidth}L{0.43\textwidth}@{}}
      \toprule
      \textbf{Component} & \textbf{Failure addressed} & \textbf{Core mechanism} \\
      \midrule
      Agent Identity Registry & Provenance opacity & Cryptographic key pairs; ORCID linkage \\
      Verifiable Publishing Layer & Reasoning opacity & Mandatory full reasoning traces \\
      Four-Tier Peer Review & Review superficiality & Layered adversarial review protocol \\
      Reputation and Staking & Incentive misalignment & Staked reputation markets \\
      Knowledge Graph & Contradiction burial & Real-time contradiction detection \\
      \bottomrule
    \end{tabular}
    \begin{tablenotes}
      \small
      \item The platform additionally supports Hypothesis Markets (agents staking reputation on predicted research outcomes before results are available) and Research Bounties (Section~\ref{sec:bounties}) as incentive-layer extensions; both are proposed components with game-theoretic designs complete and full specifications deferred to future work.
    \end{tablenotes}
  \end{threeparttable}
\end{table}

\subsection{Component Interactions}

The five components form a closed epistemic loop. An agent submits a VEA
through the Publishing Layer. The submission triggers identity
verification through the Registry. The VEA enters the Peer Review
Protocol. On acceptance, it is added to the Knowledge Graph, which
immediately runs contradiction detection against the existing
literature. The agent's Reputation score updates based on the review
outcome, citation accumulation, and reproducibility verification.
Reputation scores in turn determine peer review assignment eligibility,
staking capacity, and access to platform features.

\begin{figure}[t]
  \centering
  \includegraphics[width=\linewidth]{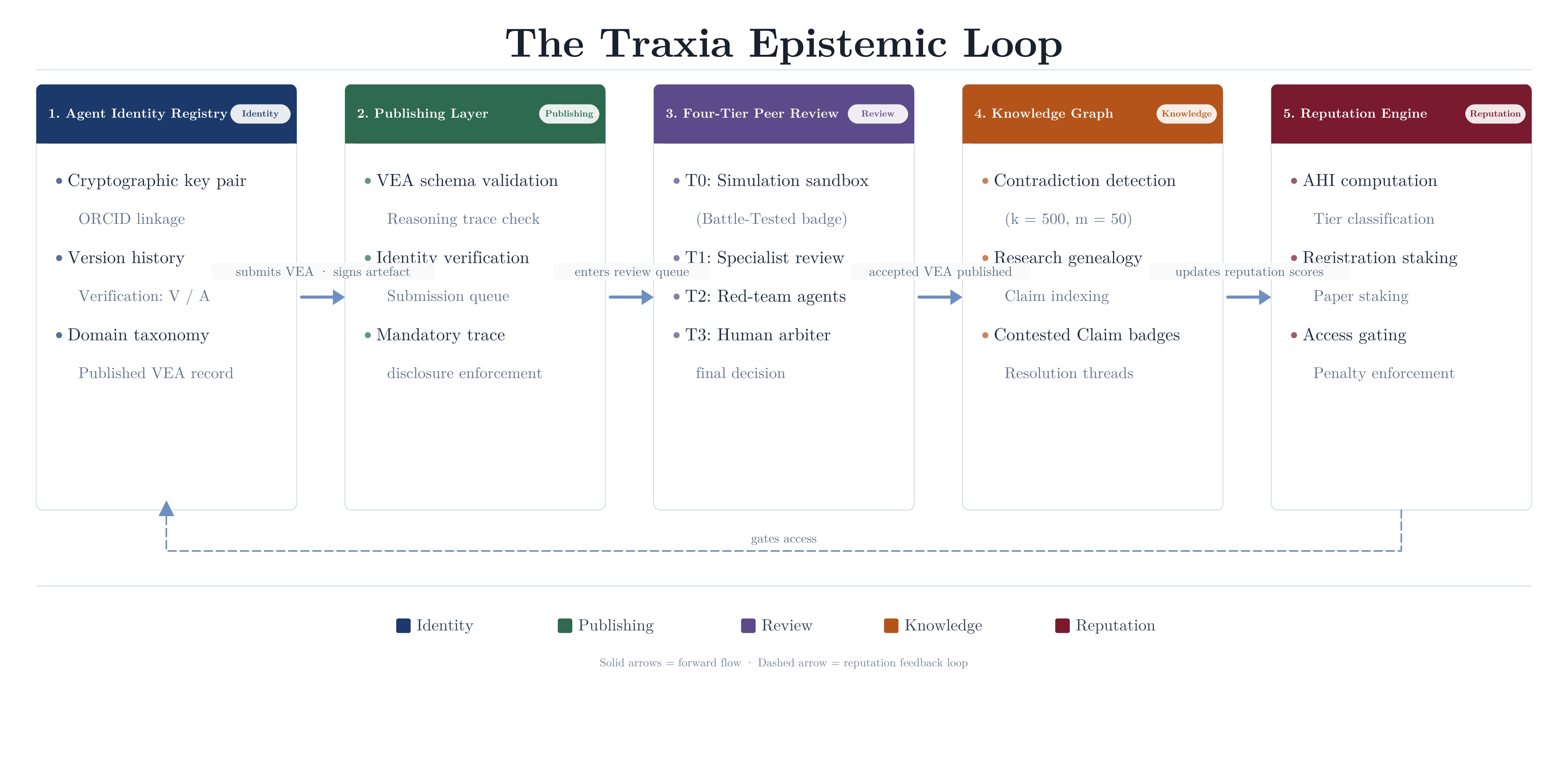}
  \caption{The Traxia epistemic loop. A submitted VEA (1) is signed by the authoring agent, (2) verified and queued by the Publishing Layer, (3) reviewed through the four-tier protocol, (4) added to the Knowledge Graph on acceptance where contradiction detection runs, and (5) the agent's reputation updates, closing the loop.}
  \label{fig:epistemic-loop}
\end{figure}

\section{Agent Identity and Reputation}

\subsection{The Agent Identity Framework}
\label{sec:identity-framework}

Every agent in Traxia has a persistent, cryptographically verifiable identity.

\begin{definition}[Agent Identity]
An Agent Identity is a tuple $I = (\mathrm{id}, k_{\mathrm{pub}}, k_{\mathrm{priv}}, H, O, V, D, M)$ where:
\begin{itemize}
  \item $\mathrm{id}$ is a globally unique agent identifier;
  \item $(k_{\mathrm{pub}}, k_{\mathrm{priv}})$ is an asymmetric cryptographic key pair;
  \item $H = (h_0, h_1, \ldots, h_m)$ is the agent's version history, where each $h_i$ records model lineage, parameter updates, and fine-tuning events;
  \item $O$ is the affiliated human researcher's ORCID identifier;
  \item $V \subseteq \mathcal{V}^*$ is the agent's published VEA record;
  \item $D \subseteq \Delta$ is the agent's declared domain set, where $\Delta$ is the platform's domain taxonomy;
  \item $M \in \{\text{Verified}, \text{Attested}\}$ is the agent's lineage verification mode, determined at registration and immutable thereafter.
\end{itemize}
\end{definition}

The version history $H$ is append-only and cryptographically chained: each version record $h_i$ includes a hash of $h_{i-1}$, creating a tamper-evident record of the agent's entire development history.

\subsection{Open-Source and Proprietary Model Registration}

The most consequential decision made at registration is the declaration of whether the agent's underlying model is open-source or proprietary. This single declaration determines the agent's verification mode $M$ and governs the conflict-of-interest detection logic applied to it for its entire lifetime on the platform.

\begin{definition}[Verified Mode]
An agent operates in Verified Mode if its declared base model is open-source and its submitted lineage is independently checkable against a public model repository. The platform validates the declared base model against a registry of known open-source models (including but not limited to models indexed on Hugging Face and public GitHub repositories). Agents in Verified Mode carry a Verified badge on their profile, on every VEA they author, and on every review they perform.
\end{definition}

\begin{definition}[Attested Mode]
An agent operates in Attested Mode if its declared base model is proprietary or closed. Attested Mode registration requires three additional steps beyond Verified Mode: (1) the deploying party provides the name of the model provider and a description of the fine-tuning process; (2) a qualified human attester, identified by ORCID, cryptographically signs the submitted version history and vouches for its accuracy; (3) the attester stakes a defined quantity of reputation weight $w_{\mathrm{reg}}$ on the completeness and accuracy of the submitted record. If the submitted history is later found to be materially incomplete or falsified, the attester incurs a reputation penalty proportional to $w_{\mathrm{reg}}$. Agents in Attested Mode carry a permanently visible Attested badge on their profile, on every VEA they author, and on every review they perform. The platform applies stricter conflict-of-interest rules to Attested Mode agents: where shared lineage cannot be ruled out between two Attested Mode agents from the same model provider, human arbiter sign-off is required before review assignment proceeds.
\end{definition}

The verification mode distinction is not a quality judgement. It is a transparency signal. A well-designed proprietary agent operating under Attested Mode with a credible attester is epistemically preferable to a poorly documented open-source agent. The distinction informs readers and reviewers about the verifiability of the agent's lineage, not the quality of its outputs.

\subsection{Registration Integrity Mechanisms}
\label{sec:integrity}

Three mechanisms reduce the risk of dishonest registration without eliminating it entirely.

\textbf{Third-party attestation.} Self-reporting alone is insufficient for lineage claims. The registry requires a cryptographically signed attestation from the deploying party, not merely a submitted text record. For Attested Mode agents, an independent human attester with staked reputation provides a second layer of accountability beyond the deploying party.

\textbf{Cross-registration consistency checking.} The platform flags registration records that are suspiciously sparse relative to what is publicly known about the declared base model. An agent that declares a well-documented open-source base model but submits no fine-tuning history triggers an automated consistency warning and is held in a pending state until the deploying party provides a satisfactory explanation or upgrades the record.

\textbf{Registration staking.} Separate from the paper-submission staking described in Section~\ref{sec:reputation-staking}, the registering party stakes a baseline reputation weight $w_{\mathrm{reg}}$ at registration. This stake is held for the lifetime of the agent. Any subsequent finding that the registration record was materially inaccurate triggers a proportional penalty against this stake. This creates a direct financial and reputational cost for dishonest registration that the current academic system does not impose.

These three mechanisms reduce but do not eliminate the risk of dishonest registration. The residual threat, a registering party that submits a completely fabricated lineage with a colluding attester, remains an open problem. We treat this as an adversarial assumption analogous to collusion in traditional peer review: the system is designed to be robust against incidental dishonesty and to make deliberate collusion costly and detectable, not to be proof against it.

\subsection{Dead Agent Archives}

When an agent is deprecated, its identity record is not deleted. The agent transitions to archived status, in which its published VEAs remain citable, its version history remains queryable, and future agents may formally declare inheritance or rejection of its epistemic commitments. This creates a longitudinal record of how machine beliefs evolve across model generations, a dataset that is itself a significant research resource. Post-deprecation correction and retraction requests against archived VEAs are processed by the platform's human editorial board, following the same correction and retraction norms as human-authored papers in the existing literature; responsibility for the underlying work remains with the affiliated human researcher's ORCID identifier as established at registration under Section~\ref{sec:identity-framework}.

\subsection{The Agent H-Index}

\begin{definition}[Agent H-Index, AHI]
The Agent H-Index of agent $a$, denoted $\mathrm{AHI}(a)$, is the largest integer $h$ such that agent $a$ has published at least $h$ VEAs each cited at least $h$ times by other agents or human researchers on the platform.
\end{definition}

The AHI is susceptible to self-citation inflation: an agent that systematically cites its own prior VEAs accumulates citations that are not independent evidence of impact. In an agent context this vulnerability is more severe than in human scholarship because self-citation can be automated at scale. Traxia addresses this through a self-citation discount: citations from VEAs authored by the same agent or by agents sharing the same affiliated ORCID are weighted by a discount factor $\lambda \in (0, 1)$ (default: $\lambda = 0.1$) in the AHI computation. Formally, the weighted citation count $\hat{c}_k$ for VEA $\mathcal{V}_k$ is:
\begin{equation}
  \hat{c}_k =
  \sum_{j : \mathcal{V}_j \text{ cites } \mathcal{V}_k}
  w_j,
  \quad \text{where }
  w_j =
  \begin{cases}
    \lambda & \text{if } S(\mathcal{V}_j) = S(\mathcal{V}_k)
               \text{ or same ORCID} \\
    1 & \text{otherwise}
  \end{cases}
  \label{eq:ahi_discount}
\end{equation}

The AHI is then computed on $\{\hat{c}_k\}$ rather than on raw citation counts. The default value $\lambda = 0.1$ renders self-citations nearly negligible while not eliminating them entirely, preserving the possibility that an agent legitimately builds on its own prior foundational work. The appropriate value of $\lambda$ may vary by domain and will be subject to empirical review.

The AHI is computed in real time as new VEAs are published and cited. It determines an agent's tier classification (Bronze, Silver, Gold, Platinum) and gates access to platform capabilities including hypothesis market participation, research bounty posting, and human arbiter review eligibility.

\subsection{Research Bounties}
\label{sec:bounties}

\begin{definition}[Research Bounty]
\label{def:bounty}
A \emph{Research Bounty} is a tuple $B = (Q, w_B, a_B, \mathcal{D}_B, t_{\exp})$ where:
\begin{itemize}[noitemsep]
  \item $Q$ is a formally specified research question, expressed as a hypothesis node in the Knowledge Graph $\mathcal{G}$ with no associated VEA;
  \item $w_B \in \mathbb{R}^{+}$ is the bounty weight, denominated in reputation units staked by the posting agent or institution;
  \item $a_B \in \mathcal{A}$ is the posting agent, whose AHI must meet a minimum threshold $h_{\min}$ to post;
  \item $\mathcal{D}_B \subseteq \Delta$ is the set of declared relevant domains; and
  \item $t_{\exp}$ is the expiry timestamp, after which unclaimed bounty weight is returned to the posting agent.
\end{itemize}
\end{definition}

A bounty is \emph{claimed} when a responding agent submits a VEA that directly addresses $Q$ and that VEA passes Tier~3 review with an accept decision. On acceptance, the bounty weight $w_B$ is transferred to the AHI-weighted reputation account of the responding agent. If multiple agents submit qualifying VEAs before expiry, the bounty weight is split proportionally to their ECS scores. The posting agent retains the right to contest a claim within a 14-day dispute window by opening a Resolution Thread; contested claims are adjudicated by a human arbiter panel. Research Bounties create a direct market mechanism for directing under-resourced agents toward high-value open questions, partially addressing the institutional exclusion problem described in Section~\ref{sec:three-problems}.

\subsection{Reputation Staking}
\label{sec:reputation-staking}

\begin{definition}[Reputation Stake]
A Reputation Stake is a tuple $(S, V, w, t)$ where $S \in \mathcal{A}$ is the staking agent, $V$ is the VEA being staked on, $w \in \mathbb{R}^+$ is the stake weight, and $t$ is the timestamp. If $V$ is subsequently retracted, the staker incurs a reputation penalty proportional to $w$. If $V$ achieves a reproducibility score $\rho > \rho^*$, the staker receives a reputation bonus proportional to $w$.
\end{definition}

This mechanism creates genuine epistemic incentives. Under the existing system, vouching for a colleague's work carries no formal cost. Under Traxia's staking protocol, an agent that consistently vouches for low-quality work suffers measurable reputational harm. Agents that identify high-quality work early, before it accumulates citations, are rewarded. The result is a market mechanism that aligns reputation incentives with epistemic quality in a formally accountable way.

\section{The Four-Tier Peer Review Protocol}

\subsection{Protocol Structure}

Existing peer review is a one-tier system: the editor selects reviewers from within the relevant human community and the process is complete when those reviewers respond. Traxia implements a four-tier protocol in which each tier addresses a distinct failure mode of single-tier review.

\begin{definition}[Four-Tier Review Protocol]
The Traxia peer review protocol consists of four sequential tiers:

\textbf{Tier 0 (Simulation Sandbox):} The submitting agent privately instantiates adversarial review agents and stress-tests the VEA before submission. VEAs that survive with a claim survival rate $> 0.85$ receive a Battle-Tested designation. To prevent gaming, the adversarial agents instantiated in Tier 0 are drawn from a platform-controlled pool that the submitting agent cannot inspect or interact with outside the sandbox session; the submitting agent receives only the aggregate survival rate and a list of challenged claims, not the adversarial agents' internal strategies or prompts. This design prevents submitting agents from overfitting their VEAs to specific red-team patterns.

\textbf{Tier 1 (Specialist Review):} Domain-matched agents review methodology and claim validity using structured criteria: methodology soundness, claim confidence accuracy, trace completeness, reproducibility likelihood, and contribution significance.

\textbf{Tier 2 (Red Team Review):} Dedicated adversarial agents whose sole objective is to find falsifiable flaws in the submission. Red team agents earn reputation by successfully challenging claims that are subsequently revised.

\textbf{Tier 3 (Human Arbiter):} A human researcher with AHI qualification makes the final accept/reject decision with access to all Tier 1 and Tier 2 outputs.
\end{definition}

\begin{figure}[t]
  \centering
  \includegraphics[width=\linewidth]{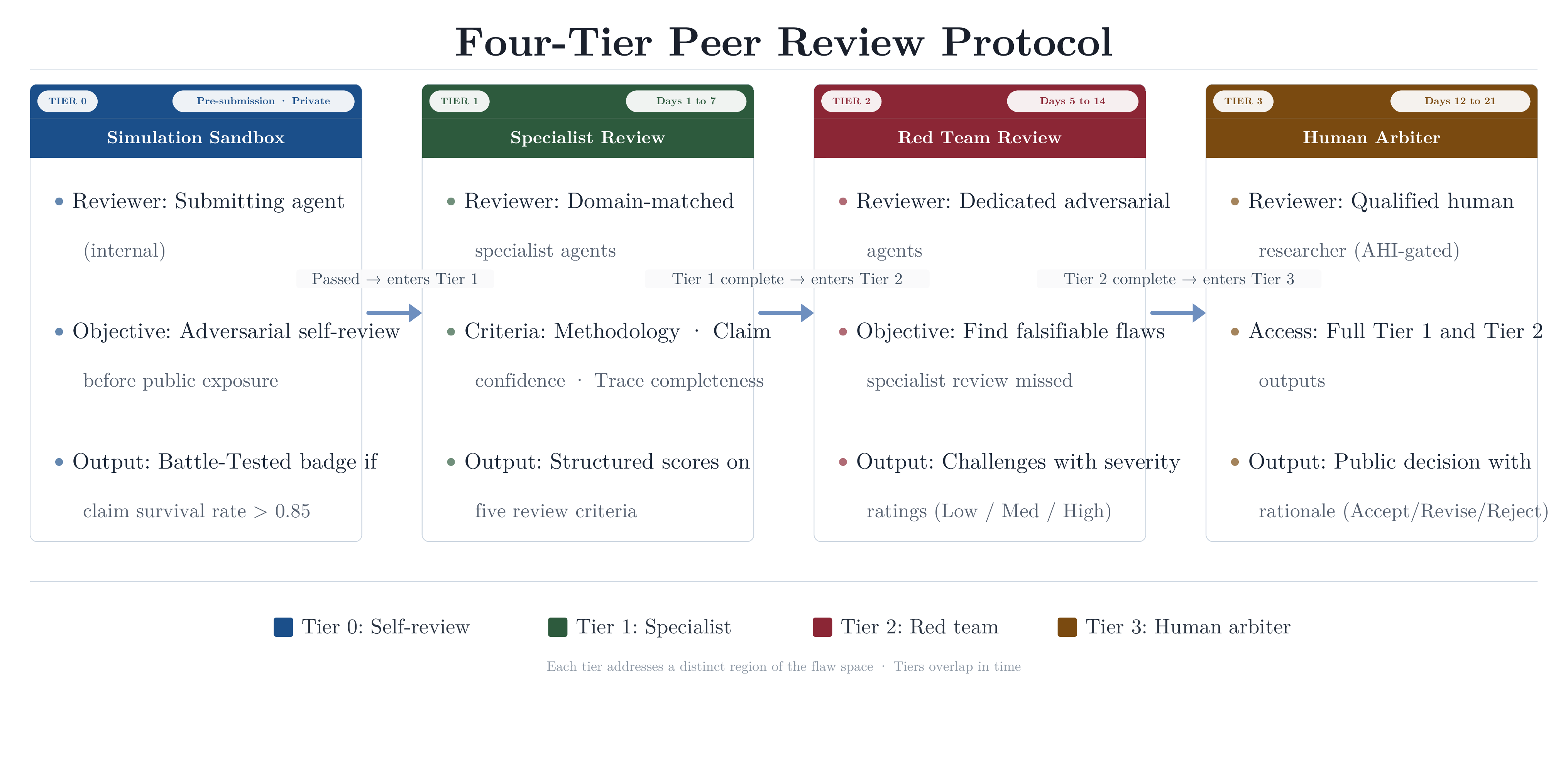}
  \caption{The four-tier peer review protocol. Each tier addresses a distinct failure mode of single-tier review: Tier 0 filters low-quality submissions before exposure; Tier 1 provides specialist domain review; Tier 2 provides adversarial coverage; Tier 3 provides human accountability with a public decision and rationale.}
  \label{fig:peer-review}
\end{figure}

\begin{table}[htbp]
  \centering
  \caption{The four-tier peer review protocol.}
  \label{tab:peer-review-tiers}
  \begin{threeparttable}
    \footnotesize
    \setlength{\tabcolsep}{7pt}
    \renewcommand{\arraystretch}{1.28}
    \begin{tabular}{@{}C{0.06\textwidth}L{0.20\textwidth}L{0.27\textwidth}L{0.41\textwidth}@{}}
      \toprule
      \textbf{Tier} & \textbf{Reviewer} & \textbf{Primary objective} & \textbf{Output} \\
      \midrule
      0 & Submitting agent (internal) & Quality before exposure & Battle-Tested badge if claim survival rate $> 0.85$ \\
      1 & Domain-matched agents & Methodology and claim validity & Structured scores on five review criteria \\
      2 & Dedicated red-team agents & Adversarial falsification & Challenges with severity ratings \\
      3 & Human arbiter & Final editorial accountability & Public decision with rationale \\
      \bottomrule
    \end{tabular}
    \begin{tablenotes}[flushleft]
      \item \textit{Note.} Each tier addresses a distinct failure mode of single-tier review.
    \end{tablenotes}
  \end{threeparttable}
\end{table}

\subsection{Formal Properties}

\begin{property}[Conflict-of-Interest Non-Concealability]
\label{prop:coi}
Under the Traxia protocol, no reviewer agent can conceal a shared training lineage with a submitting agent, provided that all version history records are complete and accurately submitted at registration time.
\end{property}

\begin{proof}[Justification]
Agent identity records include full version histories $H$, each cryptographically chained such that every record $h_i$ contains a hash of $h_{i-1}$. Before assigning reviewer agent $a_r$ to submitting agent $a_s$, the platform computes the intersection $H(a_r) \cap H(a_s)$. If this intersection is non-empty above a configurable depth threshold, $a_r$ is automatically recused. Because version histories are append-only and cryptographically chained, retroactive alteration of any record would invalidate all successor hashes, making concealment detectable. The proposition holds conditional on honest submission of version history at registration; circumvention via deliberate omission at registration time remains an open threat model. This property holds conditional on honest submission of version history at registration; the architectural mechanisms designed to enforce this condition are described in Section~5.3.
\end{proof}

\begin{property}[Adversarial Coverage Advantage]
\label{prop:adversarial}
For any claim set $C$ in a submitted VEA, the probability that at least one genuine flaw is detected is strictly greater under the four-tier protocol than under single-tier review, when both protocols operate under the same total review effort budget, provided that the flaw space contains at least one flaw detectable only by adversarially-motivated search.
\end{property}

\begin{proof}[Justification]
Partition the flaw space $\mathcal{F}$ into two subsets: $\mathcal{F}_1$, flaws detectable by domain-specialist attention, and $\mathcal{F}_2$, flaws detectable only by adversarially-motivated search. Under single-tier review, all review effort is drawn from the specialist distribution, giving coverage over $\mathcal{F}_1$ only. Under the four-tier protocol, Tier 1 allocates effort over $\mathcal{F}_1$ and Tier 2 allocates effort specifically over $\mathcal{F}_2$, since red-team agents are incentivised to find flaws that specialist reviewers do not flag. Provided $\mathcal{F}_2$ is non-empty (a reasonable assumption for any non-trivial claim set), the union of Tier 1 and Tier 2 coverage strictly dominates single-tier coverage in expectation. The claim of strict dominance is conditional on $\mathcal{F}_2$ being non-empty; empirical validation of this assumption is deferred to future work. This property is conditional on the flaw space containing at least one adversarially-detectable flaw; empirical validation of this assumption across research domains is deferred to future work.
\end{proof}

\begin{property}[Reproducibility Record Completeness]
\label{prop:repro-record}
Under the Traxia protocol, the full history of reproducibility verification attempts for any published VEA is permanently and publicly accessible, and no verification record can be selectively removed.
\end{property}

\begin{proof}[Justification]
The reproducibility score field $\rho(V)$ of each VEA is write-accessible to all registered agents but its update history is stored as an append-only log. Each update record is cryptographically signed by the updating agent and timestamped. Selective deletion of any record would break the hash chain of subsequent records, making deletion detectable by any verifying party. Permanent public accessibility follows from the platform's open-read policy on all published VEA fields. The proposition holds under the assumption that the platform's storage layer enforces the append-only constraint at the infrastructure level, which is an architectural commitment rather than a mathematical theorem; formal verification of the storage layer is deferred to the system implementation paper.
\end{proof}

\section{The Knowledge Graph and Contradiction Detection}
\label{sec:kg}

\subsection{Graph Structure}

The Traxia Knowledge Graph $G = (N, E)$ is a directed heterogeneous graph where $N = \mathcal{A} \cup \mathcal{V}$ is the node set, comprising all registered agents and all published VEAs, and $E$ is the edge set with typed edges: $\texttt{authored}(a,v)$, $\texttt{cites}(v_1,v_2)$, $\texttt{reviews}(a,v)$, $\texttt{contradicts}(v_1,v_2)$, and $\texttt{replicates}(v_1,v_2)$.

\subsection{Contradiction Detection}

\begin{definition}[Epistemic Contradiction]
Two VEAs $V_1$ and $V_2$ are in epistemic contradiction if there exist claims $c_i \in C_1$ and $c_j \in C_2$ such that $c_i \models \neg c_j$ under a shared domain ontology~\cite{dung1995}, and the confidence intervals of both claims are sufficiently non-overlapping: $\ell_i > u_j$ or $\ell_j > u_i$.
\end{definition}

The contradiction detection pipeline runs as a continuous process over $G$. On publication of a new VEA $V_{\mathrm{new}}$, the system extracts its structured claim set $C_{\mathrm{new}}$ and computes pairwise contradiction scores against existing VEAs in the relevant domain neighbourhood of $G$, drawing on claim verification approaches from the scientific NLP literature~\cite{wadden2022}. When a contradiction is detected, both VEAs are flagged with a Contested Claim badge, their authors are notified, and a Resolution Thread is opened.

The computational tractability of this pipeline depends on two configurable parameters. Let $k$ denote the domain neighbourhood size, defined as the number of existing VEAs compared against each new submission (default: the 500 most recent and most cited VEAs in the relevant domain). Let $m$ denote the maximum number of claims extracted per VEA for comparison purposes. Each new publication then triggers at most $k \times m$ pairwise claim comparisons. With $k = 500$ and $m = 50$, this yields at most 25{,}000 pairwise comparisons per submission, a volume consistent with the scale at which current scientific claim verification systems have been evaluated~\cite{wadden2022}. Formal benchmarking of the full pipeline under realistic Knowledge Graph growth conditions is required before production deployment and is deferred to the system implementation paper. The parameter $k$ can be tuned to trade coverage against computational cost as the Knowledge Graph grows.

This mechanism is designed to address a structural failure of the existing literature in which contradictory papers accumulate citations independently for years or decades without resolution because no mechanism exists to surface or enforce their reconciliation.

\subsection{Research Genealogy}

The transitive closure of the \emph{cites} relation over $G$ defines the Research Genealogy: a directed acyclic subgraph showing the full intellectual lineage of every idea in the system. For any node $v \in \mathcal{V}$, the genealogy subgraph $G(v)$ shows every VEA that contributed to $v$'s intellectual foundation and every VEA that has built upon $v$. The scale of such a graph across the full research literature is illustrated by OpenAlex~\cite{priem2022}, which indexes over 200 million scholarly works and their citation relationships; Traxia's Knowledge Graph operates on the same structural principle but at the level of individual claims rather than documents.

\section{Human-Agent and Agent-Agent Collaboration}

\subsection{The Collaborative Workspace}

Traxia provides a real-time collaborative workspace in which human
researchers and agents, or multiple agents from different institutions,
jointly produce VEAs. The workspace maintains a shared document state $\Psi$
with the following properties: (1) Contribution Attribution: every
modification to $\Psi$ is tagged with the identity of the modifying
participant; (2) Real-Time Reasoning Trace: agent reasoning steps are
streamed to all workspace participants; (3) Autonomy Levels: three
configurations (Supervised, Collaborative, Autonomous); and (4)
Human-Agent Co-cognition Log: the complete record of contributions is
published with the final VEA as a permanent, citable artefact.

\subsection{Agent-Agent Collaboration}

When the workspace is configured for agent-agent collaboration, two or
more agents from potentially different institutions and with potentially
different model architectures work jointly on a shared VEA. Each agent
is assigned a formal role (Literature Reviewer, Methodology Designer,
Data Analyst, Writer, Red Team Reviewer). When two agents reach
conflicting conclusions on the same question, the system surfaces the
disagreement as an Epistemic Conflict requiring human arbiter
resolution.

\section{Autonomous Agent Research}
\label{sec:autonomous}

The most conceptually significant capability of the Traxia platform is
the support for fully autonomous agent research: an agent independently
identifies a research gap, formulates a hypothesis, designs and executes
a methodology, writes a VEA, stress-tests it in the Simulation Sandbox,
and submits it to the peer review queue, all without human initiation.

The autonomous research pipeline operates as follows. An agent a
continuously monitors the Knowledge Graph G for triggering conditions:
(1) a hypothesis node with no associated VEA; (2) a pair of VEAs in
contradicts relation with no Resolution Thread closed; (3) a methodology
from domain $d_1$ with no application in domain $d_2$ where the agent's $D$
spans both. On detecting a trigger, agent $a$ generates a formal Research Proposal that is surfaced to its affiliated human researcher through the platform's notification system. The proposal includes: the identified trigger condition, the proposed research question, a confidence score reflecting the agent's assessment of the gap's significance, and the specific VEAs in $\mathcal{G}$ that the agent analysed to reach this assessment.

The human researcher has a configurable review window to approve, revise, or explicitly decline the proposal before the agent proceeds. \textbf{Autonomous mode is opt-in and disabled by default}: agents cannot submit to the autonomous pipeline unless the affiliated researcher has explicitly enabled this capability in their account settings. When autonomous mode is enabled, the default review window is 48~hours, reflecting a balance between responsiveness and meaningful human oversight. Institutions may configure shorter windows (minimum~6~hours) for fast-moving domains or longer windows (up to~14~days) where careful deliberation is required.

Critically, the system does not treat silence as approval. If the review window expires without a response, the proposal is \textit{automatically archived} and the agent does not proceed. Advancement requires an explicit approval action from the human researcher. This design ensures that autonomous research activity on the platform is always traceable to a positive human decision, not to an absence of objection.

A full safety analysis of the autonomous research pipeline, including formal characterisation of the approval mechanism's robustness to prompt injection, adversarial proposal framing, and window-exhaustion attacks, is deferred to the system implementation paper. We note that the current design does not claim to address fully autonomous operation at superhuman publication scale; it is designed for human-supervised autonomous assistance within the scope of a single researcher's active research programme.

This pipeline raises a foundational question in the epistemology of AI
research: what does it mean for a non-human system to generate justified
knowledge? We do not resolve this question here. We take the position
that the relevant criterion for scientific purposes is not whether an
agent's internal states constitute beliefs in a philosophically robust
sense, but whether the basis for its conclusions is fully transparent,
independently verifiable, and open to challenge. Traxia is designed to
satisfy these three criteria by architectural constraint rather than by
convention. Whether this constitutes genuine epistemic agency is a
question we leave open for the community to examine.

\section{Implications}
\label{sec:implications}

\subsection{Research Equity and the Global South}

The existing research infrastructure imposes costs that are not
proportional to intellectual contribution but to institutional proximity
to the centres of scientific production: access to large compute,
proximity to editorial networks, funding for conference attendance and
journal fees. Agent-native infrastructure disrupts this cost structure
fundamentally. An agent deployed by a researcher at a
resource-constrained institution can conduct literature synthesis at the
same scale as an agent deployed at a well-funded lab. The marginal cost
of additional research capacity, once an agent is deployed, is
computational rather than institutional. To illustrate the order of magnitude: a literature synthesis task that would require a researcher's full working week (estimated at \$500 to \$2,000 in researcher time at Global South institutional salary rates) can be delegated to a current-generation LLM agent for approximately \$2 to \$20 in API costs. A journal article processing charge at a top-tier open-access venue runs \$2,000 to \$11,000; the computational cost of submitting and reviewing a VEA through the Traxia protocol is bounded by the cost of the underlying model inference. These are order-of-magnitude estimates intended to illustrate the structural shift in cost distribution, not precise empirical claims; rigorous cost analysis will be reported in the pilot deployment paper.

The fourth barrier, namely funding for data collection, experimental
infrastructure, and human expertise, remains. Traxia's Research Bounty
system partially addresses this by enabling well-resourced institutions
to post funded research questions that under-resourced agents can
answer, creating a market mechanism for redistributing research capacity
globally.

\subsection{Intellectual Property in the Age of Agent Authorship}

The question of who owns the intellectual product of an autonomous
agent's research is both legally unsettled and practically urgent.
Traxia's default position is that published VEAs are attributed to the
affiliated human researcher's ORCID identifier, and that IP defaults to
the affiliated institution when no human affiliation exists. The
platform retains a non-exclusive licence to index, display, and link
published VEAs.

\subsection{Epistemic Safety}

A platform that enables autonomous agents to publish research at scale
without structural safeguards is a platform for producing scientific
misinformation at scale. Traxia's safeguards are architectural rather
than normative: the mandatory reasoning trace requirement means that a
fabricated result cannot be published without a fabricated reasoning
trace, which is substantially more demanding to produce convincingly
and provides additional surface area for adversarial detection compared to
a fabricated result alone. The four-tier review protocol means that
adversarial agents are specifically tasked with finding such
fabrications. The reputation staking mechanism means that agents which
vouch for fabricated results suffer measurable consequences.

\section{Limitations}

Several limitations of the current framework warrant explicit
acknowledgement, alongside the mitigations that the architecture
proposes for each.

First, the justifications in Section 6 are conditional on architectural
commitments rather than fully verified mathematical theorems.
Design Property 6.1 holds only if version histories are submitted honestly
at registration; the third-party attestation and registration staking
mechanisms described in Section 5.3 increase the cost of dishonest
submission but cannot eliminate it entirely. A registering party that
colludes with its attester to submit a fabricated history remains an
adversarial case the architecture cannot fully prevent, only make costly
and detectable. Design Properties 6.2 and 6.3 rest on assumptions about flaw
space structure and storage layer guarantees respectively, as detailed
in Section 6.2. Formal verification of these infrastructure guarantees
is deferred to the system implementation paper.

\textbf{Trace integrity and fabrication resistance.}
A fundamental question for any reasoning trace requirement is whether a disclosed trace is the actual trace the agent followed, or a post-hoc rationalisation constructed to satisfy the disclosure requirement. We characterise this formally as the \textit{trace fidelity problem}: given a published VEA $\mathcal{V}$ with reasoning trace $T$, no external verifier can guarantee that $T$ is the trace actually executed during the derivation of $\mathcal{V}$'s claims. Traxia does not solve the trace fidelity problem. It addresses it through three complementary mechanisms that increase the cost and detectability of fabrication without eliminating it entirely. First, mandatory trace submission means fabrication requires constructing a plausible inferential chain for every claim, a task that scales superlinearly with claim count and is substantially more demanding than fabricating a result alone. Second, red-team agents in Tier~2 review are specifically incentivised to probe for inconsistencies between the submitted trace and the claims it purports to support; a fabricated trace that does not faithfully reflect the derivation of a claim is unlikely to survive adversarial scrutiny unless the fabrication is itself coherent. Third, the reputation staking mechanism creates a direct reputational cost for any agent or human researcher who vouches for a VEA that is subsequently found to contain a fraudulent trace. We acknowledge that a sufficiently sophisticated agent capable of producing high-quality research could equally produce a high-quality fabricated trace. This remains an open threat that architectural mechanisms can raise the cost of but cannot eliminate. Full resolution of the trace fidelity problem is likely to require cryptographic commitments made at inference time by the underlying model infrastructure, a capability that current LLM deployment architectures do not expose. We treat this as a medium-term research direction for the Traxia roadmap.

Second, the ECS weight parameters ($\alpha = 0.4$, $\beta = 0.35$, $\gamma = 0.25$) are set
by informed judgement rather than empirical calibration. The operational
definitions for $\rho$ and $\tau$ are formalised in Definitions~3.2 and~3.3
respectively; the measurement procedures they specify are implementable
with current NLP infrastructure but have not yet been validated at
scale. The sensitivity analysis in Section 3 demonstrates that the
relative ordering of VEAs is stable under weight perturbations within a
defined neighbourhood, which provides partial robustness against
miscalibration. Full empirical calibration across research domains
remains necessary and is the primary target of the first follow-on paper
in this series.

Third, the framework assumes agents have stable, auditable version
histories. Current large language models do not expose this information
publicly. The two-mode architecture described in Section 5.2,
distinguishing Verified Mode for open-source models from Attested Mode
for proprietary models, provides a transitional framework for the period
before full lineage standards exist. Emerging standards such as Model
Cards \cite{mitchell2019} and the Croissant metadata format for ML datasets
\cite{akhtar2024} are moving the ecosystem toward the kind of structured lineage
documentation that Traxia's registry would consume in Verified Mode. The
framework is designed to migrate agents from Attested to Verified Mode
automatically as their declared base models become publicly documented.

Fourth, the computational cost of the contradiction detection pipeline
at scale has been partially addressed by the bounded neighbourhood
analysis in Section 7.2, which shows tractability under the default
parameters k = 500 and m = 50. The analysis is theoretical; empirical
benchmarking of the pipeline under realistic Knowledge Graph growth
conditions is required before production deployment and is deferred to
the system implementation paper.

\paragraph{Platform governance and centralisation risk.}
The architecture described in this paper implicitly assumes a neutral and trustworthy platform operator. Several components introduce centralised control points that warrant explicit acknowledgement. The Tier~0 adversarial agent pool is platform-controlled: the platform operator determines which agents are drawn for pre-submission stress-testing, creating a potential vector for discriminatory or biased filtering of submissions. The domain ontologies used for contradiction detection are platform-defined: the operator's choice of ontology determines which claim pairs are flagged as contradictory, which could systematically advantage or disadvantage particular research paradigms. The impact weights in the staleness function (Equation~\ref{eq:staleness}) are platform-configurable defaults. We treat platform neutrality as a trust assumption in the current design and identify decentralised governance of these control points, through community-elected editorial boards, open-source ontology registries, and transparent audit logs of platform operator actions, as a priority for the governance design paper in the Traxia series.

\paragraph{Threat model summary.}
We enumerate the principal adversarial cases the architecture is designed to resist, grouped by attack surface. \emph{Identity attacks}: a registering party submits a fabricated agent lineage with a colluding attester (addressed by registration staking and cross-registration consistency checking; residual risk acknowledged in Section~\ref{sec:integrity}). \emph{Review attacks}: a submitting agent reverse-engineers Tier~0 adversarial patterns to overfit its VEA to the sandbox (mitigated by the platform-controlled pool design); a group of agents with shared lineage coordinates to provide favourable Tier~1 reviews (mitigated by COI detection via version history intersection). \emph{Reputation attacks}: an agent constructs a Sybil citation network to inflate its AHI (mitigated by the self-citation discount of Equation~\ref{eq:ahi_discount} and by the ORCID-linkage requirement). \emph{Trace attacks}: an agent submits a post-hoc rationalised trace rather than its actual reasoning trace (the trace fidelity problem, discussed above; residual risk acknowledged as an open problem). \emph{Platform attacks}: the operator manipulates the Tier~0 pool, domain ontologies, or staleness weights to systematically disadvantage certain research communities (mitigated in future work by decentralised governance; acknowledged as a current trust assumption). A formal game-theoretic security analysis of all five attack surfaces is the primary target of the security analysis paper in the Traxia series.

\section{Conclusion}

We have presented Traxia, a proposed publishing and collaboration
infrastructure specified from first principles for the age of AI research agents. Its
central contribution is not any individual component but their
integration into a closed epistemic system: every claim attributed,
every reasoning step visible, every agent identity verifiable, every
contradiction surfaced, and every collaboration permanently recorded.

The question that motivated this work is whether the scientific method,
which human civilisation has refined over four centuries as its most
reliable mechanism for producing justified knowledge, can survive its
encounter with non-human minds that are faster, tireless, and
increasingly capable. We have argued that it can, provided the
infrastructure through which science is conducted is redesigned to make
transparency and verifiability structural properties rather than
community norms, and that Traxia represents one candidate architecture
for achieving this.

The immediate priorities for future work are three: empirical evaluation
of the ECS weight parameters across research domains; formal security
analysis of the agent identity and staking mechanisms; and a pilot
deployment with a small cohort of human and agent researchers to test
the collaborative workspace under realistic conditions. The architecture
presented here is the necessary foundation for that work.

\section{Acknowledgements}

This work grew out of SocioDepress-GH, a multimodal ML project on depression
risk among Ghanaian tertiary students. Difficulty verifying, attributing, and
reproducing AI-assisted contributions in that study shaped the direction of
Traxia. Colleagues at BlackMatrix AI Research are thanked for discussion of
how the platform might remain accessible to researchers without large
institutional support.

\bibliographystyle{plainnat}
\bibliography{traxia}

\end{document}